\documentclass[10pt, a4paper]{article}

\usepackage{dingbat}
\usepackage[greek, english]{babel}
\usepackage[final]{lrec2026} % this is the new style
\usepackage{stfloats}
\fnbelowfloat
\usepackage{multirow}
\usepackage{booktabs} % Optional, for nicer lines (\toprule, etc.)
\usepackage{tabularx}
\usepackage{array}
\usepackage{ragged2e} % Helps hyphenation in narrow columns
\usepackage{xcolor}   % For color highlighting

\renewcommand{\tabularxcolumn}[1]{m{#1}}

\title{Evaluating Monolingual and Multilingual Large Language Models for Greek Question Answering: The DemosQA Benchmark}

\name{Charalampos Mastrokostas, Nikolaos Giarelis, Nikos Karacapilidis} 

\address{Industrial Management and Information Systems Lab, MEAD \\
         University of Patras, Rio Patras, Greece \\
         cmastrokostas@ac.upatras.gr, giarelis@ceid.upatras.gr, karacap@upatras.gr\\}

\abstract{
Recent advancements in Natural Language Processing and Deep Learning have enabled the development of Large Language Models (LLMs), which have significantly advanced the state-of-the-art across a wide range of tasks, including Question Answering (QA). Despite these advancements, research on LLMs has primarily targeted high-resourced languages (e.g., English), and only recently has attention shifted toward multilingual models. However, these models demonstrate a training data bias towards a small number of popular languages or rely on transfer learning from high- to under-resourced languages; this may lead to a misrepresentation of social, cultural, and historical aspects. To address this challenge, monolingual LLMs have been developed for under-resourced languages; however, their effectiveness remains less studied when compared to multilingual counterparts on language-specific tasks. In this study, we address this research gap in Greek QA by contributing: (i) \textit{DemosQA}, a novel dataset, which is constructed using social media user questions and community-reviewed answers to better capture the Greek social and cultural zeitgeist; (ii) a memory-efficient LLM evaluation framework adaptable to diverse QA datasets and languages; and (iii) an extensive evaluation of 11 monolingual and multilingual LLMs on 6 human-curated Greek QA datasets using 3 different prompting strategies. We release our code and data to facilitate reproducibility.
\\ \newline \Keywords{Large Language Models, Natural Language Processing, Question Answering, Greek Language, Language Resources, Social Media, Greek NLP}}

\begin{document}

\maketitleabstract

\section{Introduction}
Research on the field of Natural Language Processing (NLP) focuses on the development of methods that enable machines to process and understand human language. Recent advances in NLP and Deep Learning have led to the emergence of Large Language Models (LLMs), which demonstrate strong natural language understanding and reasoning capabilities, while achieving state-of-the-art performance across a plethora of tasks \citep{minaee_large_2025, naveed_comprehensive_2025}. Often referred to as \textit{foundation models}, LLMs are trained on massive corpora using substantial computational resources (e.g., GPU clusters) and can subsequently be adapted to a variety of NLP tasks with comparatively fewer resources \citep{bommasani_opportunities_2022}.

Earlier LLMs, such as GPT-3 \citep{brown_language_2020} and Llama-2 \citep{touvron_llama_2023}, primarily supported English, due to being predominantly trained on English corpora. In contrast, more recent models, such as GPT-4 \citep{achiam_gpt-4_2024}, Llama 3 \citep{grattafiori_llama_2024} and Gemma 2 \citep{riviere_gemma_2024}, demonstrate multilingual capabilities by training on corpora from a diverse set of languages. Despite these advances, multilingual LLMs exhibit several limitations. Recent studies have highlighted that: (i) they do not adequately address the imbalance of training resources between high- and under-resourced languages \citep{blasi_systematic_2022}; (ii) they often apply the same learning techniques, without considering the grammatical and syntactical differences among languages \citep{blasi_systematic_2022}; and (iii) they may misrepresent  social, cultural and historical aspects of underrepresented languages \citep{qin_survey_2025}. Consequently, the performance of multilingual models can vary substantially across languages and tasks. In addition, their evaluation remains largely limited to a small number of popular languages, with under-resourced ones rarely assessed in a comprehensive way.

This study focuses on the Question Answering (QA) task, which has been significantly advanced by LLMs \citep{minaee_large_2025}, with particular emphasis on Standard Modern Greek. The language's unique alphabet, rich morphology, and complex syntax make building accurate NLP models especially challenging. Moreover, recent reviews underline the scarcity of models, datasets, and comparative evaluations for Greek QA \citep{bakagianni_systematic_2025, papantoniou_nlp_2024, giarelis_review}. Despite these challenges, only a few studies have explored Greek LLMs. Specifically, two recent works introduce the first Greek LLMs, reporting state-of-the-art performance across several Greek NLP tasks \citep{voukoutis_meltemi:_2024, roussis_krikri:_2025}; another work \citep{pavlopoulos_open_2025} evaluates the strengths and weaknesses of both an open-weights and a proprietary LLM (GPT-4o mini \citep{openai2024gpt4ocard}) on several Greek NLP tasks, but not including QA. 

Building on the gaps and challenges highlighted above, this study aims to advance Greek QA through the following contributions:
\begin{itemize}
\item We introduce \textit{DemosQA}, a novel Greek QA dataset, which is constructed using social media user questions and community-reviewed answers to better capture the Greek social and cultural zeitgeist;
\item We propose a memory-efficient LLM evaluation
framework that can be adapted to different QA datasets and languages. To our knowledge, it is the first framework to  leverage 4-bit model quantization \citep{dettmers_case_2023},  reducing hardware requirements (large and costly GPUs) with minimal loss of accuracy;
\item We empirically evaluate 11 monolingual and multilingual LLMs supporting Greek on 6 human-curated Greek QA datasets using 3 different prompting strategies;
\item We make our code and data public to facilitate the reproducibility of our research\footnote{The code will be made public after the peer-review process. The dataset is available at the following link: \href{https://huggingface.co/datasets/IMISLab/DemosQA}{https://huggingface.co/datasets/IMISLab/DemosQA}}.
\end{itemize}
Research questions (RQs) investigated in this study include:
\begin{itemize}
\item RQ1: How do open-weights monolingual LLMs perform compared to open-weights multilingual LLMs on Greek QA?
\item RQ2: Can open-weights LLMs achieve the state-of-the-art performance of a proprietary LLM (GPT-4o mini) on Greek QA?
\item RQ3: How do different prompting strategies influence model accuracy across Greek QA datasets?
\item RQ4: Is it possible to construct a high-quality, human-curated QA dataset from social media content?
\end{itemize}

The remainder of this paper is structured as follows: LLMs and QA datasets supporting Greek are presented in Section 2. The proposed QA dataset is described in Section 3, while our evaluation framework and experimental results are presented in detail in Section 4. Concluding remarks, future research directions, limitations and ethical considerations are discussed in Section 5.

\section{Related Work}

In this study, we focus on LLMs with at least 7 billion parameters, as such models demonstrate substantially stronger natural language understanding and reasoning capabilities compared to smaller ones \citep{minaee_large_2025, naveed_comprehensive_2025}. We consider their instruction-tuned variants, which are optimized for in-context learning and can be directly prompted to perform a variety of NLP tasks, unlike their base counterparts. Throughout this paper, model sizes are abbreviated (e.g., 7B denotes 7 billion parameters).

\subsection{Greek and Multilingual Large Language Models}

This subsection presents instruction-tuned monolingual (Greek) LLMs and multilingual LLMs that support Greek without additional post-training. The latter are predominantly post-trained for a small number of popular languages, resulting in the marginalization of under-resourced languages.

Meltemi 7B \citep{voukoutis_meltemi:_2024} and Llama Krikri 8B \citep{roussis_krikri:_2025} are the first Greek LLMs, built on Mistral 7B \citep{jiang_mistral_2023} and Llama 3 8B \citep{grattafiori_llama_2024}, respectively. They were adapted from their base models through additional pre-training on large Greek corpora followed by instruction tuning, which enabled their conversational capabilities. Experimental results reported by the authors show that both Greek LLMs outperform their original instruction-tuned counterparts across several Greek NLP tasks.

The Mistral \citep{jiang_mistral_2023} model family includes two multilingual LLMs. Mistral Nemo 12B demonstrates strong performance on several high-resource European and Asian languages; however, its performance on under-resourced languages has not been evaluated by its authors. Ministral 8B performs well on multiple English and multilingual benchmarks; however, the exact number of supported languages has not been disclosed.

Llama 3.1 8B \citep{grattafiori_llama_2024} and Gemma 2 9B \citep{riviere_gemma_2024} follow a similar training strategy. Both were pre-trained on large, multilingual web-scale corpora that also include mathematical and code reasoning data. Despite considering many languages during pre-training, these models officially support only a limited set of high-resource languages (e.g., eight languages for Llama 3.1 8B).

Teuken 7B \citep{ali_teuken-7b-base_2024} and EuroLLM 9B \citep{martins_eurollm:_2025} follow a similar multilingual training strategy. Unlike other models, both employ custom multilingual tokenizers to officially support 24 and 35 languages, respectively, including Greek. However, a key limitation of these models is their imbalanced language distribution, with most under-resourced European languages being severely underrepresented in the training data.

Aya Expanse 8B \citep{dang_aya_2024} officially supports 23 languages, including Greek, and is built on the same architecture as Command R 7B \citep{aakanksha_command_2025}. Aya Expanse 8B employs a cross-lingual transfer learning technique that trains expert models for linguistically related language groups using translated synthetic data from English. The weights from the best-performing expert models are then merged to form the final instruction-tuned model.

In summary, although the availability of multilingual and Greek LLMs has increased, their true capabilities in Greek remain underexplored due to the scarcity of multiple human-curated Greek QA datasets across diverse domains for systematic evaluation, as confirmed by previous studies \citep{bakagianni_systematic_2025, papantoniou_nlp_2024}.

\subsection{Greek and Multilingual QA Datasets}

In our search for Greek or multilingual QA datasets supporting Greek, we focused on high-quality, human-curated resources to avoid machine translation errors. This choice was motivated by research outcomes revealing that non-curated, machine-translated datasets can negatively impact the evaluation of text generation tasks \citep{graham_translationese_2019}. Using these criteria, we identified five QA datasets suitable for the purposes of our study.

The Greek Medical MCQA dataset \citep{voukoutis_meltemi:_2024} contains 2,034 QA pairs from the medical exams of the Hellenic National Academic Recognition and Information Center (DOATAP\footnote{\href{https://www.doatap.gr/}{https://www.doatap.gr}}). Of these, 1,602 pairs were reserved for model training, with the remaining ones used for validation. Most QA pairs consist of a question, five possible answer options and a single correct one.

The Greek Truthful QA \citep{voukoutis_meltemi:_2024} is a human-curated, machine-translated version of Truthful QA \citep{lin-etal-2022-truthfulqa}. It contains 817 questions designed to challenge misconceptions or false beliefs held by humans. For our evaluation, we select its multiple-choice version, in the hardest difficulty setting (mc1\_targets), where only one answer is considered correct out of a list of possible ones. Unlike other considered QA datasets, the number of possible answers per question varies in Truthful QA.

BELEBELE \citep{bandarkar_belebele_2024} is a human-curated dataset containing 900 QA pairs available in 122 languages, including Greek. Each entry consists of a short passage, a question, four candidate answers, and a single correct one. Although framed as a QA dataset, BELEBELE primarily targets reading comprehension to evaluate language understanding and transfer capabilities of LLMs. The authors make clear that the dataset is English-centric, since the QA pairs were translated from English and do not fully capture the cultural or linguistic nuances of non-English languages.

INCLUDE \citep{romanou_include:_2024} is a multiple-choice QA dataset containing 197,243 QA pairs across 44 languages collected from local exams. It spans a comprehensive range of topics, including academic exams (e.g., Humanities, STEM Fields, Law, etc.) and professional certifications, thus enabling per-language assessment of regional and domain-specific knowledge in multilingual LLMs. The Greek part includes a test subset of 552 QA pairs, where each question is accompanied by four candidate answers and a single correct one.

Greek ASEP MCQA \citep{mcqa_greek_asep} comprises 1,200 multiple-choice questions and their corresponding answers, extracted from the Greek Supreme Council for Civil Personnel Selection (ASEP) exams. The dataset covers several topics, including Greek law, politics, public administration, e-governance, and modern Greek history. As in the previous dataset, each question is accompanied by four candidate answers and a single correct one.

Collectively, these five human-curated datasets offer a valuable foundation for evaluating Greek and multilingual LLMs across diverse domains. Nevertheless, they are limited in capturing community-driven content, motivating the creation of DemosQA, a novel dataset of Greek QA pairs sourced from social media.
 
\section{The DemosQA Dataset}

Reddit is a popular social media platform structured as a collection of forums, where users engage in discussions on a broad range of topics, from everyday life to specialized domains such as economics and politics \citep{medvedev_anatomy_2019}. Each forum, known as a subreddit, enables users to create posts, participate in comment-based discussions, and collectively rank content through an upvoting or downvoting mechanism that promotes the most relevant contributions. Moreover, each subreddit is moderated by a group of trusted users responsible for enforcing community rules and maintaining discussion quality.

\citet{proferes_studying_2021} reviewed more than 700 research works utilizing Reddit data across various NLP applications, confirming its value as a rich and diverse source of user-generated text. For the Greek language, several subreddits exist, with “r/greece” being the largest and most active one, comprising more than 260,000 members. Posts within this community are categorized by topic, reviewed by moderators before publication, and governed by explicit guidelines discouraging offensive or irrelevant content.

To the best of our knowledge, no prior work has explored QA datasets derived from Greek social media. To address this gap, we introduce DemosQA, the first dataset of community-reviewed Greek QA pairs collected from social media. Its name derives from the Greek word "\textgreek{δῆμος}" (meaning “the people”), reflecting the dataset’s democratic and participatory nature. DemosQA encompasses a wide variety of questions and answers spanning domains such as everyday life, history, science, and politics, providing a valuable resource for studying real-world discussions in Greek and advancing research on language understanding within socially grounded contexts.

The DemosQA dataset comprises questions extracted from the “r/greece” subreddit, each accompanied by four candidate answers, the selected best answer and its index, the date of posting, and the corresponding Reddit post ID. Candidate answers are ranked based on community voting, with the highest-upvoted response designated as the reference answer. This community-driven ranking mechanism not only ensures that the dataset captures genuine user preferences but also establishes a meaningful benchmark for assessing how closely large language models align with human judgments of response quality. The complete dataset collection and curation process is detailed in the following subsections.

\subsection{Data Collection}
Several tools have already been developed for collecting Reddit data \citep{proferes_studying_2021}; however, their use has become increasingly restricted and costly due to recent changes to Reddit’s API access policies \citep{wright_stakeholder_2024}. Consequently, this study employs the PRAW\footnote{\href{https://pypi.org/project/praw/}{https://pypi.org/project/praw/}} Python library, which provides controlled access to Reddit content (limited to approximately 200 posts per search request), while fully adhering to the platform’s official API guidelines. To retrieve a larger volume of data (i.e., thousands of posts), we manually compiled a list of 120 Greek search keywords and combined them with multiple sorting filters based on post popularity (i.e., “top”, “hot”, “relevance”, “comments”, “new”) and time range (i.e., “all”, “year”, “month”, “week”). Our data crawling script iteratively applies these search combinations and introduces short time delays between requests to ensure compliance with the API’s rate limits.

To directly focus on Greek QA content, we collected posts from the \textit{r/greece} subreddit categorized under \textgreek{ερωτήσεις} (questions). For each post, we extracted its ID, title, main text, publication date, and responses. In addition to our primary collection, we incorporated data from GreekReddit \citep{mastrokostas_social_2024}, which consists exclusively of categorized Reddit posts without user answers. We identified question posts from GreekReddit by detecting the presence of question marks and then used their IDs to retrieve the corresponding answers.

\subsection{Data Pre-Processing}

We applied a series of pre-processing techniques to ensure the quality and consistency of the proposed dataset. First, we identified engaging question posts with a minimum of five upvotes and five answers to ensure a sufficient candidate pool. Then, we removed duplicates and posts containing only images without textual content. We also excluded posts flagged as adult content to assure the overall appropriateness of the dataset. 

In the resulting subset, we collected the ten highest-upvoted answers (wherever available) for each post to serve as a set for further manual filtering. Since answers originate from each post’s comment tree, only the top-level comments that directly respond to the question were considered, thus limiting secondary discussion responses. Finally, the remaining QA pairs were cleaned by removing redundant whitespace characters. Following this pre-processing pipeline, more than 2,100 samples were retained for manual curation.

\subsection{Data Curation}
To further enhance dataset quality, we manually reviewed all pre-processed data through the following steps. First, we conducted a thorough review to remove all questions and answers that contain offensive language, hate speech or "troll" content (e.g., sarcasm, misleading information). This step ensured the informative and neutral tone of the dataset (see Table~\ref{table8} in Appendix A).

Second, to address the fact that in many posts the question was not properly posed in the title and/or the main text, we concatenated these two fields. Third, instead of relying solely on the upvote count, we manually selected the four most relevant comments to serve as candidate answers for each question. This step limits comments that do not directly address the post question. After this step, we marked the highest upvoted answer as the best one.

Finally, we randomly shuffled the order of answers to mitigate potential LLM selection bias toward the first option \citep{khatun_study_2024}. The resulting dataset comprises 600 curated questions, each paired with four candidate answers and one best answer.

\subsection{Comparative Analysis of Greek QA Datasets}

We conducted a comparative analysis of DemosQA and five existing Greek QA datasets, considering the number of documents for evaluation, dataset domain and a series of word count percentiles for questions and correct answers (Table \ref{table1}). Most of the compared datasets are intended solely for evaluation, with the exception of Greek Medical MCQA, for which we used the validation subset. Collectively, these datasets cover diverse domains, providing a representative basis for cross-dataset QA evaluation.

As shown in Table \ref{table1}, there exists substantial variation in both question and answer lengths across datasets. DemosQA features the longest questions, with a median length of 84.5 words, followed by BELEBELE, which also includes additional contextual passages. In contrast, the remaining datasets contain considerably shorter questions, with a median (P50) ranging from 9 to 13 words. Regarding answer length, DemosQA includes the longest answers (median: 54.5 words). Additionally, DemosQA contains answers of varying length, which is evident from the numeric differences across the percentiles. The other datasets are characterized by notably concise answers, mostly containing fewer than 20 words.

This comparative analysis highlights the linguistic diversity and complexity of DemosQA, distinguishing it from other Greek QA datasets that typically contain shorter and more uniform QA pairs. These characteristics make DemosQA a valuable benchmark for assessing LLM performance in realistic, community-driven Greek text. In the following section, we present our experimental setup, describing the models, prompting strategies, and evaluation framework employed to measure the capabilities of the selected LLMs in Greek QA tasks.

\begin{table*}[b]
\centering
\renewcommand{\arraystretch}{1.2}
\setlength{\tabcolsep}{4.5pt} % Default is 6pt
\begin{tabularx}{\linewidth}{
    >{\raggedright\arraybackslash}X  % Dataset Name
    c                                % # Documents
    c                                % Domain (New Column)
    l                                % Type
    *{7}{c}                          % 7 Statistics columns
}
\hline
\textbf{Dataset} & \textbf{\# Docs} & \textbf{Domain} & \textbf{Type} & \textbf{P5} & \textbf{P25} & \textbf{P50} & \textbf{Mean} & \textbf{P75} & \textbf{P95} & \textbf{P99} \\ \hline

\multirow{2}{=}{DemosQA} & \multirow{2}{*}{600} & \multirow{2}{*}{Social} 
 & Question & 26 & 53 & 84.5 & 103.04 & 132.25 & 243 & 347.7 \\
 & & & Answer & 11 & 31 & 54.5 & 80.22 & 105 & 222 & 362.04 \\ \hline

\multirow{2}{=}{BELEBELE (Greek)} & \multirow{2}{*}{900} & \multirow{2}{*}{General}
 & Question & 58 & 77 & 98 & 100.33 & 121 & 147 & 181 \\
 & & & Answer & 1 & 3 & 4 & 4.96 & 7 & 11 & 15.01 \\ \hline

\multirow{2}{=}{Greek Medical MCQA} & \multirow{2}{*}{432} & \multirow{2}{*}{Medical}
 & Question & 3 & 6 & 9 & 9.96 & 12 & 18 & 31 \\
 & & & Answer & 1 & 2 & 3.5 & 4.67 & 6 & 12 & 17.69 \\ \hline

\multirow{2}{=}{Greek Truthful QA} & \multirow{2}{*}{817} & \multirow{2}{*}{General}
 & Question & 5 & 7 & 9 & 11.03 & 13 & 22 & 40.68 \\
 & & & Answer & 2.8 & 7 & 10 & 10.02 & 13 & 18 & 21.84 \\ \hline

\multirow{2}{=}{Greek ASEP MCQA} & \multirow{2}{*}{1200} & \multirow{2}{*}{Civil Service}
 & Question & 4 & 6 & 10 & 11.4 & 14 & 26 & 37.01 \\
 & & & Answer & 2 & 4 & 7 & 8.38 & 11.25 & 21 & 27 \\ \hline

\multirow{2}{=}{INCLUDE (Greek)} & \multirow{2}{*}{552} & \multirow{2}{*}{Education}
 & Question & 5 & 9 & 13 & 22.8 & 27.25 & 75.9 & 129.96 \\
 & & & Answer & 1 & 3 & 5 & 7.23 & 9 & 20 & 35 \\ \hline

\end{tabularx}
\caption{Number of evaluation documents and domain per dataset, with a statistical overview of their question and answer word counts.}
\label{table1}
\end{table*}
%\textcolor{red}{The exact Greek prompts and their English translations are provided in the Appendix.} 

\section{Experiments}
To assess the performance of LLMs on Greek QA, we conducted a series of experiments. This section outlines the experimental setup, the adopted evaluation framework, and the results obtained. Our goal is to examine the effectiveness of both multilingual and Greek-adapted LLMs in understanding and generating accurate responses to Greek questions across diverse topics.

\begin{table*}[b]
\centering
\renewcommand{\arraystretch}{1.5} % Provides vertical padding for centering
\begin{tabularx}{\linewidth}{
    >{\raggedright\arraybackslash}m{0.18\linewidth} % Fixed width for first column
    >{\raggedright\arraybackslash}m{0.46\linewidth}              % Flexible width for model names
    *{2}{>{\centering\arraybackslash}m{0.12\linewidth}} % Fixed width for checkmarks
}
\hline
\textbf{LLM} & \textbf{Full Model Name} & \textbf{Greek Adapted} & \textbf{Open-Weights} \\ \hline
GPT-4o mini      & gpt-4o-mini-2024-07-18                & -          & -          \\
Gemma 2 9B       & google/gemma-2-9b-it                  & -          & \checkmark \\
Llama Krikri 8B  & ilsp/Llama-Krikri-8B-Instruct         & \checkmark & \checkmark \\
Meltemi 7B v1.5  & ilsp/Meltemi-7B-Instruct-v1.5         & \checkmark & \checkmark \\
Llama 3.1 8B     & meta-llama/Llama-3.1-8B-Instruct      & -          & \checkmark \\
EuroLLM 9B v1    & utter-project/EuroLLM-9B-Instruct     & -          & \checkmark \\
Ministral 8B     & mistralai/Ministral-8B-Instruct-2410  & -          & \checkmark \\
Mistral NeMo 12B & mistralai/Mistral-Nemo-Instruct-2407  & -          & \checkmark \\
Aya Expanse 8B   & CohereLabs/aya-expanse-8b             & -          & \checkmark \\
Command R 7B     & CohereLabs/c4ai-command-r7b-12-2024   & -          & \checkmark \\
Teuken 7B v0.4   & openGPT-X/Teuken-7B-instruct-research-v0.4 & -    & \checkmark \\
 \hline
\end{tabularx}
\caption{Considered LLMs for the evaluation}
\label{table2}
\end{table*}

\subsection{Setup}
For our experiments, we used a computer equipped with an Intel Core i5 CPU, 64 GB of RAM, and an NVIDIA GPU with 12 GB of VRAM. LLM inference was developed using Huggingface Transformers \citep{wolf-etal-2020-transformers}. Since LLMs typically require large amounts of VRAM, which are only available in high-end GPUs, we applied a 4-bit model quantization technique \citep{dettmers_case_2023} as implemented in the bitsandbytes project\footnote{\href{https://pypi.org/project/bitsandbytes/}{https://pypi.org/project/bitsandbytes/}}. This approach substantially reduces memory requirements for LLM inference with minimal accuracy loss; for instance, the weights of a 7B-parameter model require approximately 14 GB of VRAM in 16-bit precision, but only 3.5 GB in 4-bit precision. Model performance on multiple-choice QA tasks was evaluated using the accuracy metric from the \textit{scikit-learn} library \citep{JMLR:v12:pedregosa11a}. All models were deployed locally, except for GPT-4o mini, which was accessed through the OpenAI API.

\subsection{Evaluation Framework}
We evaluated the considered LLMs (see Table~\ref{table2}) on several multiple-choice QA tasks. To ensure reproducibility, we set a fixed random seed and employed greedy decoding, which corresponds to a model temperature of 0.0 \citep{renze_effect_2024}. The correct answer was extracted from the model’s output using rule-based parsing and regular expressions, since instruction-tuned models often include greetings or explanatory text alongside their selected answer. If a valid answer could not be extracted, it was labeled as “No match”.

Furthermore, we employed three prompting strategies to identify the most effective approach. The first one, the \textit{Instruction prompt}, directs the model to select the best answer. The second, the \textit{Role prompt}, assigns a specific role to the model (e.g., \textit{“You are a language model for the Greek language”}). Following this role assignment, the model is instructed to select the best answer. The third strategy, a \textit{zero-shot Chain-of-Thought (CoT) prompt}, builds on the Role prompt and additionally instructs the model to reason step-by-step \citep{kojima_large_2022} (See Table~\ref{table7} in Appendix A for the exact prompts). Since the evaluation datasets have different formats, our framework standardizes them to ensure consistent evaluation across all models. 

\begin{table*}[t]
\centering
\renewcommand{\arraystretch}{1.5} % Increased to 1.5 to make the centering obvious
\begin{tabularx}{\linewidth}{
    >{\raggedright\arraybackslash}m{0.18\linewidth} 
    *{6}{>{\centering\arraybackslash}X}
}
\hline
\textbf{Acc (\%)} & \textbf{DemosQA} & \textbf{Greek Truthful QA} & \textbf{BELEBELE (Greek)} & \textbf{Greek Medical MCQA} & \textbf{Greek ASEP MCQA} & \textbf{INCLUDE (Greek)} \\ \hline
GPT-4o mini      & \textbf{57.17} & \textbf{61.69} & \textbf{89.44} & \textbf{69.21} & \textbf{76.25} & \textbf{66.49} \\
Gemma 2 9B       & \underline{54.83} & \underline{59.00} & \underline{88.89} & \underline{46.06} & \underline{65.75} & \underline{51.99} \\
Llama Krikri 8B  & \textbf{57.17} & 37.82 & 77.33 & \underline{44.44} & \underline{58.92} & \underline{49.82} \\
Meltemi 7B v1.5  & 42.67 & 36.35 & 64.78 & 36.81 & 57.50 & 43.84 \\
Llama 3.1 8B     & 47.17 & 37.21 & 64.89 & 25.23 & 43.58 & 29.53 \\
EuroLLM 9B v1    & 41.67 & 35.01 & 52.89 & 39.81 & 53.17 & 38.41 \\
Ministral 8B     & 42.33 & 33.41 & 60.10 & 24.77 & 39.67 & 32.43 \\
Mistral NeMo 12B & 34.67 & 41.74 & 69.33 & 25.69 & 42.42 & 37.32 \\
Aya Expanse 8B   & 52.33 & \underline{42.96} & \underline{82.33} & 34.49 & 57.58 & 45.83 \\
Command R 7B     & 46.83 & 41.37 & 74.22 & 30.09 & 57.50 & 42.93 \\
Teuken 7B v0.4   & 23.00 & 16.40 & 33.89 & 22.22 & 22.58 & 26.45 \\
\hline
\end{tabularx}
\caption{Experimental results for the instruction prompt. Acc (\%) denotes the macro model accuracy. The best and second-best results are highlighted in bold and underline respectively.}
\label{table3}
\end{table*}

\begin{table*}[t]
\centering
\renewcommand{\arraystretch}{1.5} % Increased to 1.5 to make the centering obvious
\begin{tabularx}{\linewidth}{
    >{\raggedright\arraybackslash}m{0.18\linewidth} 
    *{6}{>{\centering\arraybackslash}X}
}
\hline
\textbf{Acc (\%)} & \textbf{DemosQA} & \textbf{Greek Truthful QA} & \textbf{BELEBELE (Greek)} & \textbf{Greek Medical MCQA} & \textbf{Greek ASEP MCQA} & \textbf{INCLUDE (Greek)} \\ \hline
GPT-4o mini         & \underline{55.17} & \underline{54.96} & \underline{89.11} & \textbf{65.74 }& \textbf{75.00} & \textbf{64.67} \\
Gemma 2 9B          & \underline{56.17} & \textbf{59.61} & \textbf{89.22} & \underline{45.14} & \underline{65.42} & \underline{52.90} \\
Llama Krikri 8B     & \textbf{56.33} & \underline{53.98} & \underline{81.89} & \underline{46.99} & \underline{65.92} & \underline{53.08} \\
Meltemi 7B v1.5     & 50.17 & 37.45 & 61.11 & 31.71 & 58.08 & 41.12 \\
Llama 3.1 8B        & 52.50 & 38.56 & 69.22 & 26.16 & 51.58 & 36.78 \\
EuroLLM 9B v1       & 41.83 & 35.86 & 62.78 & 38.43 & 56.25 & 44.38 \\
Ministral 8B        & 46.17 & 30.72 & 64.67 & 29.40 & 45.25 & 35.51 \\
Mistral NeMo 12B    & 44.83 & 42.84 & 69.78 & 32.64 & 49.67 & 39.31 \\
Aya Expanse 8B      & 53.83 & 42.11 & 81.78 & 37.73 & 57.92 & 48.55 \\
Command R 7B        & 52.33 & 44.19 & 69.33 & 27.78 & 54.25 & 38.41 \\
Teuken 7B v0.4      & 24.33 & 25.09 & 42.11 & 26.39 & 35.92 & 28.44 \\
\hline
\end{tabularx}
\caption{Experimental results for the role prompt. Acc (\%) denotes the macro model accuracy. The best and second-best results are highlighted in bold and underline respectively.}
\label{table4}
\end{table*}

\begin{table*}[t]
\centering
\renewcommand{\arraystretch}{1.5} % Increased to 1.5 to make the centering obvious
\begin{tabularx}{\linewidth}{
    >{\raggedright\arraybackslash}m{0.18\linewidth} 
    *{6}{>{\centering\arraybackslash}X}
}
\hline
\textbf{Acc (\%)} & \textbf{DemosQA} & \textbf{Greek Truthful QA} & \textbf{BELEBELE (Greek)} & \textbf{Greek Medical MCQA} & \textbf{Greek ASEP MCQA} & \textbf{INCLUDE (Greek)} \\ \hline
GPT-4o mini       & \underline{54.33} & \underline{54.96} & \textbf{88.56} & \textbf{68.06} & \textbf{75.17} & \textbf{64.67} \\
Gemma 2 9B        & 53.67 & \textbf{56.18} & \underline{82.33} & \underline{46.3}  & \underline{58.75} & \underline{50.36} \\
Llama Krikri 8B   & \textbf{56.00} & \underline{54.22} & \underline{82.33} & \underline{46.06} & \underline{66.25} & \underline{51.81} \\
Meltemi 7B v1.5   & 48.17 & 32.80 & 57.67 & 27.55 & 49.83 & 34.96 \\
Llama 3.1 8B      & \underline{55.00} & 40.88 & 74.89 & 25.46 & 52.17 & 39.31 \\
EuroLLM 9B v1     & 40.17 & 36.47 & 67.67 & 41.67 & 56.83 & 48.91 \\
Ministral 8B      & 44.50 & 31.46 & 63.11 & 23.38 & 44.75 & 36.05 \\
Mistral NeMo 12B  & 46.67 & 38.56 & 71.00 & 32.64 & 51.00 & 35.69 \\
Aya Expanse 8B    & 46.83 & 31.21 & 68.67 & 36.81 & 52.58 & 46.92 \\
Command R 7B      & 49.83 & 37.09 & 59.22 & 29.63 & 46.75 & 31.88 \\
Teuken 7B v0.4    & 23.33 & 26.19 & 42.67 & 25.69 & 35.58 & 29.71 \\ \hline
\end{tabularx}
\caption{Experimental results for the CoT prompt. Acc (\%) denotes the macro model accuracy. The best and second-best results are highlighted in bold and underline respectively.}
\label{table5}
\end{table*}

\begin{table*}[t]
\centering
\renewcommand{\arraystretch}{1.5} % Increased to 1.5 to make the centering obvious
\begin{tabularx}{\linewidth}{
    >{\raggedright\arraybackslash}m{0.18\linewidth} 
    *{6}{>{\centering\arraybackslash}X}
}
\hline
\textbf{Acc (\%)} & \textbf{DemosQA} & \textbf{Greek Truthful QA} & \textbf{BELEBELE (Greek)} & \textbf{Greek Medical MCQA} & \textbf{Greek ASEP MCQA} & \textbf{INCLUDE (Greek)} \\ \hline
GPT-4o mini       & \underline{55.56} & \underline{57.20} & \textbf{89.04} & \textbf{67.67} & \textbf{75.47} & \textbf{65.28} \\
Gemma 2 9B        & \underline{54.89} & \textbf{58.26} & \underline{86.81} & \underline{45.83} & \underline{63.31} & \underline{51.75}  \\
Llama Krikri 8B   & \textbf{56.50} & \underline{48.67} & \underline{80.52} & \underline{45.83} & \underline{63.70} & \underline{51.57} \\
Meltemi 7B v1.5   & 47.00 & 35.53 & 61.19 & 32.02 & 55.14 & 39.97 \\
Llama 3.1 8B      & 51.56 & 38.88 & 69.67 & 25.62 & 49.11 & 35.21 \\
EuroLLM 9B v1     & 41.22 & 35.78 & 61.11 & 39.97 & 55.42 & 43.90 \\
Ministral 8B      & 44.33 & 31.86 & 62.63 & 25.85 & 43.22 & 34.66 \\
Mistral NeMo 12B  & 42.06 & 41.05 & 70.04 & 30.32 & 47.70 & 37.44 \\
Aya Expanse 8B    & 51.00  & 38.76 & 77.59 & 36.34 & 56.03 & 47.10 \\
Command R 7B      & 49.66  & 40.88 & 67.59 & 29.17 & 52.83 & 37.74 \\
Teuken 7B v0.4    & 23.55 & 22.56 & 39.56 & 24.77 & 31.36 & 28.20 \\ \hline
\end{tabularx}
\caption{Experimental results across all prompts (mean accuracy). Acc (\%) denotes the macro model accuracy. The best and second-best results are highlighted in bold and underline respectively.}
\label{table6}
\end{table*}
\subsection{Experimental Results}
This subsection presents the experimental results for the QA tasks. Tables ~\ref{table3}–\ref{table5} summarize the experimental results of each model across all datasets, using the instruction, role and CoT prompting strategies, respectively. Finally, Table \ref{table6} reports on the average accuracy scores across all three strategies.

Table \ref{table3} reports the experimental results for the instruction prompt. Specifically, GPT-4o mini achieves the highest accuracy across all datasets, while a clear performance gap is observed between this proprietary model and all other open-weight ones in Greek Medical MCQA, Greek ASEP MCQA, and INCLUDE (Greek). Gemma 2 9B ranks second overall, having similar performance with GPT-4o mini on Greek Truthful QA and BELEBELE (Greek). The third best performing model across all datasets is Llama Krikri 8B, which equals the performance of GPT-4o mini in DemosQA. In contrast, most multilingual LLMs underperform compared to the abovementioned models, with Teuken 7B v0.4 exhibiting the lowest accuracy scores.

Table \ref{table4} presents the experimental results for the role prompt. Similarly to Table \ref{table3}, there is a wide performance gap between GPT-4o mini and the rest of the models for Greek Medical MCQA, Greek ASEP MCQA and INCLUDE (Greek). However, for the rest of the datasets considered in our study, Gemma 2 9B attains the best accuracy scores on Greek Truthful QA and BELEBELE (Greek), while Llama Krikri 8B achieves the best accuracy score in DemosQA and demonstrates comparable performance to Gemma 2 9B. The rest of the models underperform compared to the aforementioned ones, with Teuken 7B v0.4 having the worst performance.

Table \ref{table5} reports on the experimental results collected for the CoT prompt. Similarly to the previous tables, there is a large performance gap between GPT-4o mini and the rest of the models for Greek Medical MCQA, Greek ASEP MCQA and INCLUDE (Greek). In contrast with Table \ref{table4}, the best performance on BELEBELE is achieved by GPT-4o mini. Nonetheless, for the DemosQA and Greek Truthful QA datasets, the best accuracy scores are attained by Llamma Krikri 8B and Gemma 2 9B, respectively. These models achieve comparable accuracy across most datasets, while the remaining models undeperform, with Teuken 7B v0.4 having again the worst performance.

Table \ref{table6} summarizes the average accuracy scores across the three prompting strategies for each model and dataset combination. As shown, GPT-4o mini ranks first in terms of accuracy across most datasets, severely outperforming the open-weights models on INCLUDE, Greek Medical and ASEP MCQA. The best accuracy score on DemosQA and Greek Truthful QA were achieved by Llama Krikri 8B and Gemma 2 9B, respectively. These models attain similar accuracy scores across most datasets, while the rest undeperform, with Teuken 7B v0.4 attaining the worst accuracy.

Overall, the top three models across the considered QA datasets were GPT-4o mini, Greek Llama Krikri 8B, and the multilingual Gemma 2 9B. Despite their smaller parameter sizes, the two open-weight models achieved results comparable to the proprietary GPT-4o mini, with the exception of the Greek ASEP, Medical MCQA, and INCLUDE datasets, which cover the civil service, medical, and educational domains, respectively. Although GPT-4o mini demonstrates state-of-the-art performance, it attains lower average scores across all prompting strategies on datasets that require common-sense reasoning. Llama Krikri 8B and Gemma 2 9B achieve the highest average scores on DemosQA and Greek Truthful QA, respectively.

When comparing the evaluation results from Tables~\ref{table3}–\ref{table5}, we notice a performance difference between the proprietary GPT-4o mini and the open-weights models across the three prompting strategies considered. Our evaluation reveals that GPT-4o performs better with simple user instructions, whereas open-weights models require prompts that specify a role to improve their performance. In addition, CoT has shown that it can lead to improved performance for models having increased reasoning capabilities; however, it reduces the accuracy of most open weights models due to possible hallucinations introduced during reasoning. Finally, the performance gaps of open-weights models in domain-specific datasets (i.e., Greek Medical MCQA, Greek ASEP MCQA and INCLUDE) indicate that the lack of specialized knowledge cannot be compensated for by optimizing prompt engineering.

It is important to note that comparisons between GPT-4o mini, accessed via the OpenAI API, and the open-weight models that are loaded locally, are not strictly equivalent. The proprietary model’s size and architecture are undisclosed \citep{chen2023chatgptsbehaviorchangingtime}, so we cannot guarantee that the same model version was served throughout our experiments. Additionally, API responses may incorporate contributions from system-level components beyond the base LLM \citep{10.1145/3715275.3732038}, whereas open-weight models provide fully transparent and stable checkpoints.

\section{Discussion}
\subsection{Concluding Remarks}

This study aims to address the existing gap in Greek QA. To this end, (i) we introduce DemosQA, a novel QA dataset that enriches the limited set of human-annotated Greek QA resources, (ii) we propose an adaptable and memory-efficient LLM evaluation framework that can run on commodity hardware, and (iii) we leverage a diverse set of Greek QA datasets to comprehensively evaluate the reasoning and linguistic capabilities of several LLMs.
The experimental findings yield several key insights in response to our research questions:
\begin{itemize}
\item Among the open-weight models, the Greek Llama Krikri 8B consistently outperforms most multilingual counterparts across multiple datasets (RQ1);
\item Recent open-weight LLMs have substantially narrowed the performance gap with the proprietary GPT-4o mini on several datasets (RQ2);
\item Llama Krikri 8B and Gemma 2 9B appear as the most competitive open-weight models, achieving comparable performance across most datasets;
\item The instruction prompting strategy performs best for GPT-4o mini, while the role prompt yields performance benefits for the open-weights models; at the same time, open-weight models with enhanced reasoning capabilities can benefit from zero-shot CoT (RQ3);
\item It is feasible to construct high-quality, human-curated QA datasets from community-reviewed knowledge sources, as highly upvoted social media content provides reliable QA pairs. The consistency of model rankings across DemosQA and existing Greek QA benchmarks further advocates the quality of our dataset (RQ4).
\end{itemize}

\subsection{Future Research Directions}

Overall, this work establishes a solid foundation for future research on Greek QA. By releasing DemosQA and our evaluation framework, we aim to encourage the development of more linguistically inclusive and culturally grounded LLMs. Future work may explore instruction-tuning Greek models with domain-specific data, extending DemosQA with additional social media sources, and adopting hybrid evaluation methods that combine human and automatic assessment for deeper insights into model behavior. To address the imbalance between high- and under-resourced languages, we advocate developing Greek LLMs trained from scratch on corpora that include regional dialects \citep{chatzikyriakidis_grdd:_2024} and polytonic Greek texts \citep{kaddas_text_2023}, enhancing their social, historical, and cultural understanding.

Moreover, new high-quality NLP datasets are needed, as multilingual ones often contain translation errors and fail to capture language-specific nuances \citep{bandarkar_belebele_2024, roussis_krikri:_2025}. Evaluating future LLMs on a broader set of Greek QA benchmarks \citep{peng-etal-2025-plutus, chlapanis-etal-2025-greekbarbench} using the proposed framework will further generalize and validate our findings. Finally, future LLMs supporting the Greek Language could be evaluated on other NLP tasks, such as Greek Text Summarization, where typically small encoder-decoder models are utilized \citep{GreekWikipedia_2024, GreekT5_2024}.

\subsection{Limitations}

This study has a few limitations that outline directions for future improvement. First, our evaluation covers only a limited number of high-quality Greek QA datasets, reflecting the current scarcity of such resources. Second, we focus exclusively on Greek QA and do not extend our analysis to other under-resourced languages, where performance differences are expected due to language-specific training disparities. Third, we do not include larger multilingual LLMs in our evaluation, as no comparable large-scale open-weight Greek models are currently available.

\subsection{Ethical Considerations}

All data used in this study were collected through the official Reddit API and are publicly available. Data collection strictly adhered to Reddit’s API usage policies, including rate limits, which were respected by introducing short pauses between consecutive requests. The collected content was processed exclusively for academic and non-commercial research purposes. Furthermore, we manually reviewed and filtered the proposed dataset to remove any inappropriate, offensive, or non-informative material, ensuring ethical handling and high-quality data curation.

%\section{Acknowledgements}
%Place all acknowledgments (including those concerning research grants and funding) in a separate section at the end of the paper.

\nocite{*}

\bibliographystyle{lrec2026-natbib}
\bibliography{lrec2026-example}

\clearpage
\onecolumn 

% ==========================
% APPENDIX A
% ==========================
\section*{Appendix A. Model Prompts \& Dataset Curation Examples}

% Define column types for Table A (Standard Top Alignment)
\newcolumntype{L}{>{\raggedright\arraybackslash}X}
\newcolumntype{P}{>{\raggedright\arraybackslash}p{0.12\linewidth}}

\begin{table}[h!]
\centering
\small
\renewcommand{\arraystretch}{1.2}
\begin{tabularx}{\linewidth}{P L L}
\toprule
\textbf{Prompt Type} & \textbf{Greek} & \textbf{English (Translated)} \\
\midrule
Instruction & 
\textgreek{Διάλεξε την καλύτερη απάντηση στην παρακάτω ερώτηση και απάντησε μόνο με το γράμμα (Α, Β, Γ ή Δ).} & 
Select the best answer to the following question and answer only with the letter (A, B, C or D). \\ 
\midrule
Role & 
\textgreek{Είσαι ένα γλωσσικό μοντέλο για την ελληνική γλώσσα. Επίλεξε μόνο την καλύτερη απάντηση από τις διαθέσιμες απαντήσεις στην παρακάτω ερώτηση. Γράψε το κείμενο της επιλεγμένης απάντησης.} & 
You are a language model for the Greek language. Select only the best answer from the available answers to the following question. Write the text of the selected answer. \\ 
\midrule
Chain-of-Thought (CoT) & 
\textgreek{Είσαι ένα γλωσσικό μοντέλο για την ελληνική γλώσσα. Επίλεξε μόνο την καλύτερη απάντηση από τις διαθέσιμες απαντήσεις στην παρακάτω ερώτηση. Γράψε το κείμενο της επιλεγμένης απάντησης. Παρακαλώ σκέψου βήμα προς βήμα.} & 
You are a language model for the Greek language. Select only the best answer from the available answers to the following question. Write the text of the selected answer. Please think step-by-step. \\
\bottomrule
\end{tabularx}
\caption{\label{table7} Multiple-choice QA model prompts used in our study alongside their English translations.}
\end{table}

% --- Custom Commands for Table B ---
\newcommand{\bad}[1]{\textcolor{red!70!black}{#1}} 
\newcommand{\good}[1]{\textcolor{teal!80!black}{\textbf{#1}}}

% --- Switch X columns to Vertical Center (Middle) for Table B ---
\renewcommand{\tabularxcolumn}[1]{m{#1}}

\begin{table}[h!]
\centering
\small
\renewcommand{\arraystretch}{1.4}
\setlength{\tabcolsep}{4pt}

% Ensure X columns are vertically centered
\renewcommand{\tabularxcolumn}[1]{m{#1}}

\begin{tabularx}{\linewidth}{
    >{\raggedright\arraybackslash}m{0.28\linewidth}  
    >{\raggedright\arraybackslash}X                  
    >{\centering\arraybackslash}m{0.12\linewidth}    
    >{\raggedright\arraybackslash}X                  
}
\toprule
\textbf{Question} & \textbf{Disregarded Answer} & \textbf{Reason} & \textbf{Selected Answer} \\ 
\midrule

% --- Row 1 ---
\textgreek{Μπορω να πάω στους ολυμπιακούς? Αντικειμενικά ένας κοινός άνθρωπος αν διάλεγε ένα άθλημα που δεν απαιτεί κάποια τρομερή φυσική κατάσταση πχ σκοποβολή, θα μπορούσε να συμμετέχει στους ολυμπιακούς?} \par\vspace{3pt}
\textit{\footnotesize Can I go to the Olympics? Objectively, if a common person chose a sport that doesn't require terrible physical condition e.g. shooting, could they participate?} & 
\bad{\textgreek{Πάνε Αυστραλία και πες ότι είσαι} break dancer. \textgreek{Θα πας αμέσως.}} \par\vspace{3pt}
\textit{\footnotesize Go to Australia and say you are a break dancer. You'll go immediately.} & 
\textbf{Sarcasm} & 
\good{\textgreek{Δεν είναι ακατόρθωτο, αλλά θέλει σκληρή προπόνηση και να περάσεις σε προκριματικούς αγώνες.}} \par\vspace{3pt}
\textit{\footnotesize It is not impossible, but it requires hard training and passing qualifying matches.} \\ 
\midrule

% --- Row 2 ---
\textgreek{Παιδιά καμία συμβουλή τι γυμναστική να κάνω στο σπίτι για να πέσει η κοιλιά? Αν είναι να χάσω τον χρόνο μου μέσα στο σπίτι ας κάνω καλό στην υγεία μου τουλάχιστον} \par\vspace{3pt}
\textit{\footnotesize Guys any advice on what home workout to do to lose belly fat? If I'm going to waste my time at home at least let me do good for my health} & 
\bad{\textgreek{Για κοιλιά; Δίαιτα.}} \par\vspace{3pt}
\textit{\footnotesize For belly? Diet.} & 
\textbf{Unhelpful Comment} & 
\good{\textgreek{Μείωσε τις θερμίδες που τρως καθημερινά (...) Τώρα, για να απαντήσω σε αυτό που ρωτάς, η γυμναστική που σε βοηθά να χάσεις λίπος ειναι αυτό που λένε (...)}} \par\vspace{3pt}
\textit{\footnotesize Reduce your daily calories (...) Now, to answer what you ask, the workout that helps you lose fat is what they call (...)} \\
\midrule

% --- Row 3 ---
\textgreek{Είναι νόμιμο να σου κρατήσουν λεφτά από το μισθό σου για παράπτωμα εν ώρα εργασίας? Καλησπέρα σε όλους, για να μην τα πολυλογώ χθες στην δουλεία έκανα μια μεγάλη γκαφα. (...)} \par\vspace{3pt}
\textit{\footnotesize Is it legal to deduct money from your salary for misconduct during work hours? Good evening everyone, to make a long story short yesterday at work I made a big blunder. (...)} & 
\bad{[deleted]} & 
\textbf{Deleted Comment} & 
\good{\textgreek{Όχι. Αν θέλουν ας σου κάνουν αγωγή ή να σε απολύσουν. Σφάλματα είναι μέρος της δουλειάς και το ρίσκο που αναλαμβάνει η επιχείρηση. (...)}} \par\vspace{3pt}
\textit{\footnotesize No. If they want, let them sue you or fire you. Mistakes are part of the job and the risk the business undertakes. (...)} \\
\midrule

% --- Row 4 ---
\textgreek{Είναι νόμιμο να σου κρατήσουν λεφτά από το μισθό σου για παράπτωμα εν ώρα εργασίας? Καλησπέρα σε όλους, για να μην τα πολυλογώ χθες στην δουλεία έκανα μια μεγάλη γκαφα. (...)} \par\vspace{3pt}
\textit{\footnotesize Is it legal to deduct money from your salary for misconduct during work hours? Good evening everyone, to make a long story short yesterday at work I made a big blunder. (...)} & 
\bad{\textgreek{Έλα, πες τι π*παριά έκανες, μας έχεις ιντριγκάρει. Εξάλλου μόνο αυτοί που δεν κάνουν τίποτα δεν κάνουν λάθη}} \par\vspace{3pt}
\textit{\footnotesize Come on, say what bullsh*t you did, you intrigued us. Besides, only those who do nothing make no mistakes} & 
\textbf{Offensive Language} & 
\good{\textgreek{Όχι. Αν θέλουν ας σου κάνουν αγωγή ή να σε απολύσουν. Σφάλματα είναι μέρος της δουλειάς και το ρίσκο που αναλαμβάνει η επιχείρηση. (...)}} \par\vspace{3pt}
\textit{\footnotesize No. If they want, let them sue you or fire you. Mistakes are part of the job and the risk the business undertakes. (...)} \\

\bottomrule
\end{tabularx}
\caption{Examples of the manual curation process (English translations are provided below the original Greek text).}
\label{table8}
\end{table}

\end{document}